\def\year{2020}\relax
\documentclass[letterpaper]{article} 
\usepackage{aaai20}  
\usepackage{times}  
\usepackage{helvet} 
\usepackage{courier}  
\usepackage{lipsum}
\usepackage[hyphens]{url}  
\usepackage{graphicx} 
\usepackage{graphicx}  
\usepackage{lscape}
\usepackage[]{algorithm2e}
\usepackage{adjustbox}
\usepackage{amsmath}
\usepackage{amsfonts}
\usepackage{bbm}
\usepackage{comment}
\usepackage{enumitem}
\usepackage{graphicx}
\usepackage{makecell}
\usepackage{makecell}
\usepackage[table]{xcolor}
\usepackage{rotating}
\usepackage[perpage]{footmisc}
\DeclareMathOperator*{\argmax}{arg\,max}

\usepackage[graphicx]{realboxes}
\definecolor{beaublue}{HTML}{D1E3F1}
\definecolor{beauyellow}{HTML}{F1EFD1}
\usepackage{rotating}
\nocopyright
\title{SynC: A Unified Framework for Generating Synthetic Population \\
with Gaussian Copula}

\author{Colin Wan,\textsuperscript{\rm 1} 
	Zheng Li,\textsuperscript{\rm 2,\rm 3}\thanks{Corresponding Author} 
	Alicia Guo,\textsuperscript{\rm 4}
	Yue Zhao\textsuperscript{\rm 5}\\
\textsuperscript{\rm 1}Department of Statistical Sciences, University of Toronto, Toronto, ON, Canada\\ 
\textsuperscript{\rm 2}Northeastern University, Toronto Campus, Toronto, ON, Canada\\ 
\textsuperscript{\rm 3}Arima Inc., Toronto, ON, Canada\\ 
\textsuperscript{\rm 4}PwC Canada, Toronto, ON, Canada\\
\textsuperscript{\rm 5}H. John Heinz III College, Carnegie Mellon University, Pittsburgh, PA, USA\\
colin.wan@mail.utoronto.ca, winston@arimadata.com, alicia.guo@pwc.com, zhaoy@cmu.edu
}
 
\begin{document}

\maketitle

\begin{abstract}
Synthetic population generation is the process of combining multiple socioeconomic and demographic datasets from different sources and/or granularity levels, and downscaling them to an individual level. Although it is a fundamental step for many data science tasks, an efficient and standard framework is absent. In this study, we propose a multi-stage framework called \textbf{SynC} (\textbf{Syn}thetic Population via Gaussian \textbf{C}opula) to fill the gap. SynC first removes potential outliers in the data and then fits the filtered data with a Gaussian copula model to correctly capture dependencies and marginal distributions of sampled survey data. Finally, SynC leverages predictive models to merge datasets into one and then scales them accordingly to match the marginal constraints. We make three key contributions in this work: 1) propose a novel framework for generating individual level data from aggregated data sources by combining state-of-the-art machine learning and statistical techniques, 2) demonstrate its value as a feature engineering tool, as well as an alternative to data collection in situations where gathering is difficult through two real world datasets, 3) release an easy-to-use framework implementation for reproducibility, and 4) ensure the methodology is scalable at the production level and can easily incorporate new data.
\end{abstract}

\section{Introduction}
Synthetic population is used to combine socioeconomic and demographic data from multiple sources, such as census and market research, and downscale them to an individual level. Often for privacy reasons, demographics and socioeconomical data is released at aggregated regional levels (e.g., only averages or percentages are released for a region with multiple residents). However, practitioners often find individual level data far more appealing, as aggregated data lacks information such as variances and distributions of residents within that region, and therefore an alternative to real population data is required. For the downscaled synthetic population to be useful, it needs to be fair and consistent. The first condition means that simulated data should mimic realistic distributions and correlations of the true population as closely as possible. The second condition implies that when we aggregate downscaled samples, the results need to be consistent with the original data. A more rigorous analysis is provided in the later section.

Synthetic data generation draws lots of attention from data scientists and is often seen as a way to augment training data in situations where data collection is difficult. In applications where large-scale data collection involves manual surveys (e.g., demographics), or when the collected data is highly sensitive and cannot be fully released to the public (e.g., financial or health data), synthetically generated datasets become an ideal substitute. For example, due to the Pseudoanonymisation article of the General Data Protection Regulation \shortcite{eu-269-2014}, organizations across the world are forbidden to release personally identifiable data. As a result, such datasets are often anonymized and aggregated (such as geographical aggregation, where variables are summed or averaged across a certain region). Being able to reverse-engineer the aggregation, therefore, is a key step to reconstruct the lost information.

Common techniques for synthetic population generation are synthetic reconstruction (SR) \shortcite{beckman1996creating} and combinatorial optimization (CO) \shortcite{huang2001comparison,voas2000evaluation}. Both approaches have specific data requirements and limitations which usually cannot be easily resolved. 
To address these challenges, we propose a new framework called \textbf{SynC} (\textbf{Syn}thetic Population with Gaussian \textbf{C}opula) to simulate microdata by sampling features in batches. The concept is motivated by \cite{jeong2016copula} and \cite{kao2012dependence}, which are purely based on copula and distribution fitting. The rationale behind our framework is that features can be segmented into distinct batches based on their correlations, which reduces the high dimensional problem into several sub-problems in lower dimensions. Feature dependency in high dimensions is hard to evaluate via common methods due to its complexity and computation requirements, and as such, Gaussian copulae, a family of multivariate distributions that is capable of capturing dependencies among random variables, becomes an ideal candidate for the application. 

In this study, we make the following contributions:

\begin{enumerate}
    \item We propose a novel combination framework which, to the best of our knowledge, is the first published effort to combine state-of-the-art machine learning and statistical instruments (e.g., outlier detection, Gaussian copula, and predictive models) to synthesize population data.
    \item We demonstrate SynC's value as a feature engineering tool, as well as an alternative to data collection in situations where gathering is difficult through two real world datasets in the automotive and market research industries.

    \item To foster reproducibility and transparency, all code, figures and results are openly shared\footnote{See supplementary material or \url{https://github.com/masked_due_to_double_blind_policy}}. 
    The implementation is readily accessible to be adapted for similar use cases. 
    \item We ensure the methodology is scalable at the production level and can easily incorporate new data without the need to retrain the entire model.
\end{enumerate}

\section{Related Works}

\subsection{Synthetic Reconstruction}
Synthetic reconstruction (SR) is the most commonly used technique to generate synthetic data. This approach reconstructs the desired distribution from survey data while constrained by the marginal. Simulated individuals are sampled from a joint distribution which is estimated by an iterative process to form a synthetic population. Typical iterative procedures used to estimate the joint distribution are iterative proportional fitting (IPF) and matrix ranking. The IPF algorithm fits a n-dimensional contingency table base on sampled data and fixed marginal distributions. The inner cells are then scaled to match the given marginal distribution. The process is repeated until the entries converge. 
    
IPF has many advantages like maximizing entropy, minimizing discrimination information \shortcite{ireland1968contingency} and resulting in maximum likelihood estimator of the true contingency table \shortcite{little1991models}. However, IPF is only applicable to categorical variables. The SynC framework incorporates predictive models to approximate each feature, which can be used to produce real-valued outputs as well and probability distribution that can be sampled from to produce discrete features. 
    
\subsection{Combinatorial Optimization}
 Given a subset of individual data with features of interest, the motivation behind combinatorial optimization (CO) is to find the best combination of individuals that satisfy the marginal distributions while optimizing a fitness function \shortcite{barthelemy2013synthetic}. CO is typically initialized with a random subset of individuals, and individuals are swapped with a pool of candidates iteratively to increase the fitness of the group. Compared to SR approaches, CO can reach more accurate approximations, but often at the expense of exponential growth of computational power \shortcite{wong2013optimizing}.

\subsection{Copula-Based Population Generation}
Copula is a statistical model used to understand the dependency structures among different distributions (refer to \textit{Proposed Framework} Section for details), and has been wdiely used in microsimulation tasks \shortcite{kao2012dependence}. However, downscaling is not possible, and the sampled data stay at the same level of granularity as the input. Jeong et al. discuss an enhanced version of IPF where the fitting process is done via copula functions \shortcite{jeong2016copula}. Similar to IPF, this algorithm relies on the integrity of the input data, which, as discussed before, can be problematic in real world settings. Our method, SynC, is less restrictive on the initial condition of the input dataset as it only requires aggregated level data. Therefore SynC is more accessible compared to previous approaches.

\section{Proposed Framework} \label{proposed_method}

\subsection{Problem Description}

Throughout this paper, we assume that the accessibility to $X = X_1, ..., X_M$ , where $X_i = [X_i^1, ..., X_i^d]^T$ is a $D^{th}$ dimensional vector representing input features. Each $m \in M$ represents an aggregation unit containing $n_m$ individuals. When referring to specific individuals, we use $x_{m,k}^d$ to denote the $d^{th}$ feature of the $k^{th}$ individual who belongs to the aggregation unit $m$. For every feature $d \in D$ and aggregation unit $m$, we only observe $X_m^d$ at an aggregated level, which is assumed to be an average (i.e. $X_m^d = \sum_{k=1}^{n_m} x_{m,k}^d$) in the case of numerical measurements, or a percentage (i.e. $X_m^d = \sum_{k=1}^{n_m} x_{m,k}^d/n_m$) in the case of binary measurements. We use the term \textit{coarse data} to refer to this type of observations. In applications, aggregation levels can be geopolitical regions, business units or other types of segmentation that make practical sense. 

Our proposed framework, SynC, attempts to reverse-engineer the aggregation process to reconstruct the unobserved $x_1^d, ..., x_{n_m}^d$ for each feature $d$ and aggregation level $m$. We assume that $M$ is sufficiently large, especially relative to the size of $D$, so that fitting a reasonably sophisticated statistical or machine learning model is permissible, and we also assume that the aggregation units, $n_m$, are modest in size so that not too much information is lost from aggregation and reconstruction is still possible. Thus for a $M \times D$ dimensional coarse data $X$, the reconstruction should have dimensions $N \times D$, where $N = \sum_{m=1}^{M} n_m$ is the total number of individuals across all aggregation units. This finer-level reconstruction is referred to as the \textit{individual data}.

SynC is designed to ensure the reconstruction satisfies the following three criteria \shortcite{munnich2003simulation}: 
\begin{enumerate}[label=(\roman*), leftmargin=*]
    \item For each feature, the marginal distribution of the generated individual data should agree with the intuitive distribution one expects to observe in reality.
    \item Correlations among features of the individual data need to be logical and directionally consistent with the correlations of the coarse data.
    \item Aggregating generated individual data must agree with the original coarse data for each $m \in M$ a.
\end{enumerate}

\begin{figure}[t]
\centering
    \includegraphics[width=\linewidth]{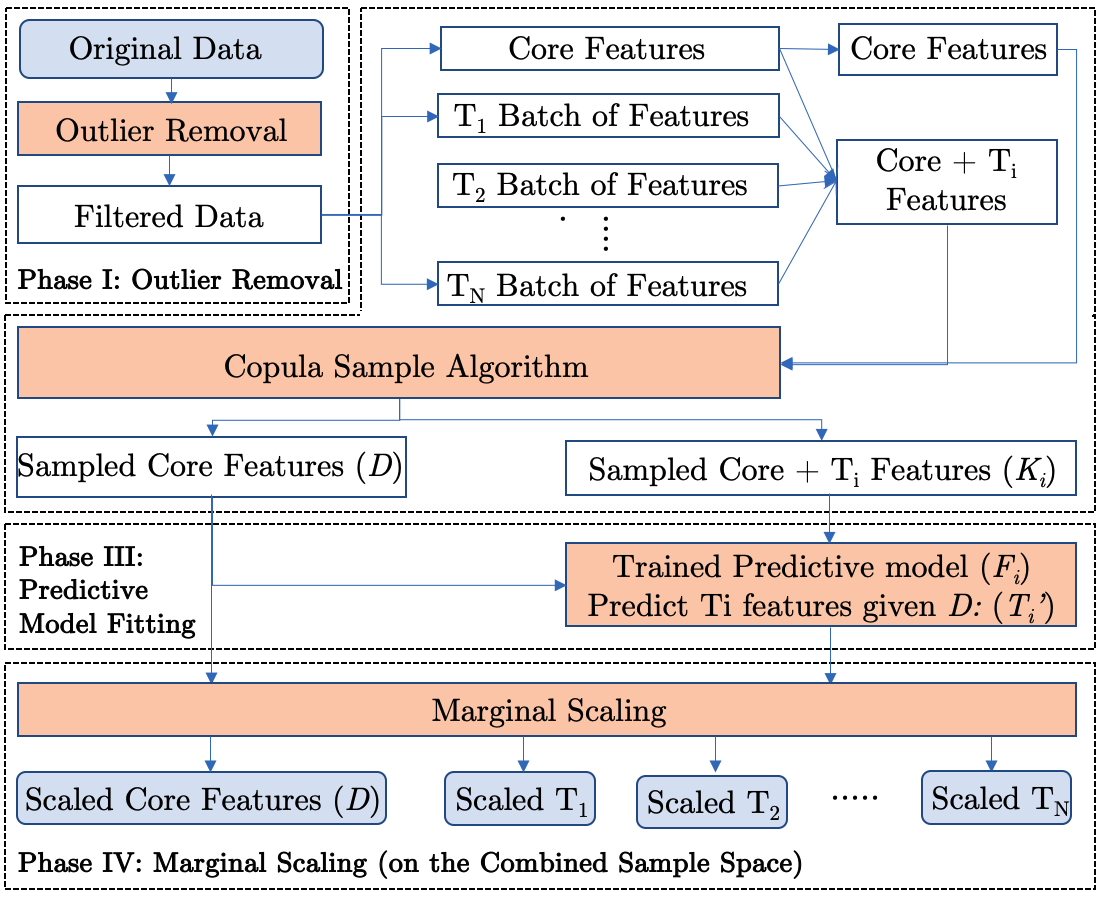}
\caption{\label{fig:flowchart}Flowchart of the SynC framework} 
\end{figure}

To illustrate the importance of these criteria, suppose that one wishes to reconstruct individual incomes from a given coarse personal incomes. The first criterion implies that the distribution of sampled individual incomes must align with the observed distribution of income, which is often modelled by the lognormal distribution. The second criterion implies that generated incomes should correlate with other money-related features such as spendings on vacations, luxury goods, and financial investments. The third criterion implies that if we average the generated incomes, we should get the input coarse personal incomes. The main idea of SynC is to simulate individuals by generating features in batches based on core characteristics and scaling the result to maintain hidden dependency and marginal constraints. As illustrated in Fig. \ref{fig:flowchart}, SynC contains four core phases and their formal description is provided below.

\subsection{Phase I: Outlier Removal} 

Outliers are the deviant samples from the general data distributions that may be a result of recording mistakes, incorrect responses (intentionally or unintentionally) or tabulation errors. The presence of outliers can lead to unpredictable results \shortcite{zhao2019pyod}. Microsimulation tasks are sensitive to outliers in the data \shortcite{passow2013adapting}, and conducting outlier detection before any analysis is therefore important. Outlier removal methods are often selected by the underlying assumptions and the actual condition of the data. When the ground truth is absent, the emerging outlier ensemble methods like SELECT \shortcite{rayana2016less} and LSCP \shortcite{zhao2019lscp} are recommended. 
    
\subsection{Phase II: Dependency Modeling} \label{Dependency Modeling}
  
We propose to use the copula model to address criteria i) and ii) since copulae, with its wide industrial applications, have been a popular multivariate modeling methodology especially when the underlying dependency is essential. First introduced by Sklar \shortcite{sklar1959fonctions}, a copula is a multivariate probability distribution where the marginal probability distribution of each variable is uniform. Let \(X=(x_1, x_2...x_D)\) be a random vector in \(R^D\), and the marginal cumulative distribution function be \(P_i(x) = Pr[x_i<x]\), define \(U_i\) as 
\small
\begin{equation}
    U = (u_1, u_2, ..., u_D) = (P_1(x_1), P_2(x_2), ..., P_D(x_D))
\end{equation}
\normalsize

A copula of the random vector \(X\) is defined as the joint CDF of a random uniform vector U:
\small
\begin{equation}
    C(u'_1, u'_2, ... u'_D) = Pr(u_1<u'_1, u_2<u'_2, ..., u_D<u'_D)
\end{equation}
\normalsize

In other words, we can describe the joint distribution of a random vector \(X\) using its marginal distributions and some copula functions. Additionally, Sklar's Theorem states that for any set of random variables with continuous CDFs, there exists a unique copula as described above. It allows us to isolate the modeling of marginal distributions from their underlying dependencies. Sampling from copulae is widely used by empiricists to produced deserved multivariate samples based on a given correlation matrix and the desired marginal distributions of each of the components. Nelsen \shortcite{nelsen2007introduction} outlines a simpler version of Algorithm \ref{algo1} for bivariate sampling, which can be generalized to multivariate cases.

\RestyleAlgo{ruled}
\begin{algorithm}[t] \label{algo1}
    \small
    \SetAlgoLined
    \KwData{Coarse Data}
    \KwResult{Simulated Individual Data}
    initialization\;
    M = number of aggregated unit\\
    D = dimension of coarse data\\
    \(\Gamma\) = $D$-dimensional correlation matrix of features of coarse data\\
    $F^{-1}_d$ =  inverse CDF of the marginal distribution of the $d^{th}$ feature\\
    \For{$m$ in $1...M$}{
    Draw $Z_{m} = Z_m^1, \cdots, Z_m^D \sim N(0,\Gamma)$, where
    $N(\mu, \Gamma)$ denotes a $d$-dimensional 
    Normal distribution with
    mean $\mu$ and correlation matrix $\Gamma$ \\
        \For{$d$ in $1...D$}{
            $u_m^d = \Phi(Z_m^d)$\\
            $y_m^d = F^{-1}_d(u_m^d)$ \\
            (This implies that $Y_m^d$ follows the desired distribution) \\
            }
        \textbf{Return} $Y_m = \{Y_m^1, ..., Y_m^D\}$ 
        }
        \textbf{Return} $Y = \{Y_1,...,Y_M\}$
    \caption{Gaussian copula sampling}
\end{algorithm}
        
In order to properly specify $F^{-1}_d$, we make a reasonable assumption that the size of each aggregation unit is significant and diverse enough such that $var(X_m^d)$ is approximately constant $\forall m$. This assumption implies that $var(X_m^d)$ can be estimated by $(\sum_{k=1}^{m} (X_m^d)^2 - \bar{X}_.^d)/M-1$, the unbiased sample variance estimator. Thus, given $\mu_m^d$ is observed from aggregation unit averages, $F^{-1}_d$ can be uniquely specified so long as we are willing to make an assumption on the distribution. For example, if the desired output is a positive continuous variable (such as income), $F_{Y_m^d}(y)$ follows a lognormal distribution
with mean $\mu_m^d$ and standard deviation $\sigma_m^d$.
In the case of categorical variables, $F_{Y_m^d}(y)$ follows a beta distribution with parameters $\alpha$ and $\beta$ such that $\frac{\alpha}{\alpha+\beta} = \mu_m^d$ and $\frac{\alpha\beta}{(\alpha+\beta)^2(\alpha+\beta+1)} = (\sigma_m^d)^2$. Our assumptions imply that $\mu_m^d$ and $\sigma_m^d$ can be derived from coarse data mean and standard deviation for feature $d$.

\small
\begin{equation}
    \mu_m^d = \textit{mean of feature $d$ in aggregation unit $m$}
\end{equation}
\begin{equation}
    \sigma_m^d = \sigma^d * \sqrt{M} * \sqrt{n_m}
\end{equation}
\normalsize
    
Algorithm 1 incorporates the above assumptions with Gaussian copula to satisfy criteria i) and ii).
    
\RestyleAlgo{ruled}
\begin{algorithm}[t]
\small
\SetAlgoLined
\KwData{Coarse Data}
\KwResult{Simulated Individual Data}
    initialization\;
    $X$ = Input coarse data\\
    $B$ = total number of batches of non-core features\\
    $S$ = predefined set of core features\\
    $T_i$ = $i^{th}$ batch of non-core features\\
    $Y$ = Initial sampled individual data with only core features using Algorithm 1\\
    \For{$i$ in $1...B$}{
        $X_i' = X[S\cup T_i]$\\
        $K_i$ = Sampled data by applying Algorithm 1 to $X'$\\
        $F(T_i \mid S = s)$ = an approximated distribution function trained on $K_i$\\
        $Q_i$ = $\argmax_{Q}$ $F(Q\mid S = Y)$\\
        $Y$ = $Y\Join Q_i$ \text{where $\Join$ is the natural join operator}
        }
    \textbf{Return} $Y$
    \caption{Batch sampling via Gaussian copula}
\end{algorithm}

\subsubsection{Gaussian Copula}
Being one of the most popular and simple form of copulae, Gaussian Copula is easy to interpret, implement and sample (and will be the copula used in section 4). It is constructed from multivariate normal distribution through probability integral transformation. For a given covariance matrix $\mathbb{R}$, Gaussian Copula requires the joint distribution of given variables can be expressed as:
    \small
    \begin{equation}
        C_{R}^{\text { Gauss }}(u)=\Phi_{R}\left(\Phi^{-1}\left(u_{1}\right), \dots, \Phi^{-1}\left(u_{d}\right)\right)
    \end{equation}
    \normalsize
    where $\Phi$ is the CDF of normal distribution. 
    \small
    \begin{equation}
        V = \left({\Phi^{-1}\left(u_{1}\right)}, \ldots, {\Phi^{-1}\left(u_{d}\right)}\right)
    \end{equation}
    \normalsize
    \small
    \begin{equation}
         c_{R}^{\mathrm{Gauss}}(u)=\frac{1}{\sqrt{\operatorname{det} R}} \exp (-\frac{1}{2}\left(V \cdot\left(R^{-1}-I\right)\cdot V^{t} \right)\\
    \end{equation} 
    \normalsize
    
    Because of the popularity of the normal distribution and the other advantages mentioned above, Gaussian copula has been the most applied copula. The earliest application of Gaussian Copula was in the finance industry: Frees and Valdez applied it in insurance \shortcite{frees1998understanding}, and Hull and White used it in credit derivative pricing \shortcite{hull2006valuing}. Recently, Gaussian Copula's application can be found in many fields such as linkage analysis \shortcite{li2006quantitative}; functional disability data modeling\shortcite{dobra2011copula}; power demand stochastic modeling \shortcite{lojowska2012stochastic}, etc. 

    \subsection{Archimedean Copula}
    After the work of Abel (1826) and Ling (1965), Archimedean Copula is introduced with the definition: 
    \small
    \begin{equation}
        C\left(u_{1}, \ldots, u_{p}\right)=\psi\left(\sum_{i=1}^{p} \psi^{-1}\left(u_{i}\right)\right)
    \end{equation}
    \normalsize
    where the $\psi :[0,1] \rightarrow[0, \infty)$ is a real valued function satisfying a few conditions \footnote{
$(-1)^{k} d^{k} \psi(x) / d^{k} x \geq 0$ for all x $\geq$ 0 and $k=1$, \ldots, $p-2$,  and 
$(-1)^{p-2} \psi^{p-2} (x)$ \text { is non-increasing and convex.}}. 

With different $\psi$ function, Archimedean Copula have many extensions. The popular cases are Clayton's Coupla which as been used in bivariate truncated data modelling \shortcite{wang2007analysis} and hazard scenarios and failure
probabilities modelling \shortcite{salvadori2016multivariate}, and Frank's Copula, which has been used in storm volume statistics analysis \cite{salvadori2004analytical} and drought frequency analysis \cite{wong2013comparison}.

    
    
    
    Archimedean Copulae have gained some attention in recent years and their unique generator functions have outstanding performances in particular situations. Yet the fitting process for the generator function can be computational expensive. As SynC preforms data generation in batches and wishes to be computational efficient, we will use Gaussian copula for modeling the dependency among features. 
    
\subsection{Phase III: Predictive Model Fitting}
In theory, one could construct a complex enough correlation matrix with all variables, and then fit a giant copula model using Algorithm 1. However, two reasons make this approach nearly impossible: 1) the curse of dimensionality makes it difficult when $D$ is sufficiently large, and 2) certain columns of $X$ may update more frequently, and it is inefficient to train new models each time. In this section, we introduce an alternative method, Algorithm 2, to resolve this issue with minimal computation requirements.

From $X$, we select, $S \in X^D$, a subset of $X$ containing the utmost important features in the given context. Next, we apply Algorithm 1 to $S$ and the resulting data set, $Y$, should satisfy the first two criteria mentioned in \textbf{Phase II}. 

One crucial question is the choice of \textit{core features}, as it is somewhat arbitrary to decide which variables should be considered core features. The choice should be made specifically for every application, as in some cases the importance and influence of some variables clearly dominate others, while in other cases quality (e.g., sampling method) or availability (e.g., sample size) of data makes some variables better candidates for inclusion in the core feature set. 
    
Secondly, divide the non-core variables into a total of $B$ batches, $T_1...T_B$, based on their correlations. Again, the division is somewhat arbitrary but in most cases, they can be based on content, method of collection or other groupings that make sense to the practitioners. Together with $S$, all $T_i$ are inputted into Algorithm 2 sequentially. The resulting sampled data set, $K_i$, contains features from $S$ and $T_i$. 
    
However, a problem immediately presents: the sampled individuals in $K_i$ do not match $Y$. Furthermore, individuals sampled in each $K_i$ does not necessarily match the sampled individuals in $K_j$ (for $j \neq i$). To fix it, we can train a predictive model on $K_i$ to approximate the distribution of $P(T_i \mid S = s)$, and use the model to predict the values of features $K_i$ given their values of the core features from $X'$. The choice of the predictive model depends on the complexity and nature of the data on a case by case basis.
    
Finally, we merge the predicted values with the original data set $Y$ and iterate until all features are covered. 
    
\subsection{Phase IV: Marginal Scaling}
The final step is to address criterion iii), which is to ensure sampled individual data agree with the input coarse data. 

If $Y^d$ is categorical with $\eta$ classes, we first note that the predicted values from \textbf{Phase III} for individual $k$ in aggregated unit $m$, represented by $\mathbf{p_{m,k}^d}=\{p_{m, k, i}^d\}_{i=0}^\eta$, is a probability distribution. Hence it is natural to assume $Y^d \sim Multi(1, \mathbf{p_{m,k}^d})$. To determine the exact class of individual $k$, we generate a random sample from the distribution. After initial sampling, the percentage of each category may not match the marginal constraint. To resolve this, SynC first randomly removes individuals from the categories that are over-sampled until their marginal constraints are satisfied. SynC then resamples the same amount of individuals using the normalized probability of the under-sampled categories. It is noted that the above processes are iterated until the desired result is achieved. 

If $Y^d$ is continuous, the sampled mean and variance should be in proximity with the original coarse data given the way $F^{-1}_d$ is constructed in \textbf{Phase II}. In case of small discrepancy, SynC will horizontally shift each data point by the difference between sample mean and the coarse data. 

\begin{table}[t]
    \resizebox{\columnwidth}{!}{  
        \begin{tabular}{c|c|c|c|c}
            \hline
            \textbf{Postal} & \# \textbf{Population} & \textbf{Avg Age} & \makecell{\textbf{\% with} \\ \textbf{Mortgage}} & 
            \makecell{\textbf{\% Speaks} \\ \textbf{two languages}} \\ 
            \hline
            \hline
            M5S3G2      & 467              & 35.1        & 0.32             & 0.69                       \\ \hline
            V3N1P5      & 269              & 37.2        & 0.35             & 0.67                       \\ \hline
            L5M6V9      & 41               & 49.1        & 0.67             & 0.43                       \\ \hline
        \end{tabular}
    }
    \caption{An excerpt of three variables from the census data}\label{tab1}
\end{table}

\begin{sidewaystable}

    \centering

    \begin{adjustbox}{max width=\textheight}
    \begin{tabular}{c|c|c|c|c|c|c|c|c|c|c}
    \hline
    \textbf{Postal} &  
    \textbf{Sex} & 
    \textbf{Age} & 
    \textbf{Ethnicity} &
    \makecell{\textbf{Immigration} \\ \textbf{Status}}  &
    \textbf{Education} &\textbf{Profession} &
    \makecell{\textbf{Marital} \\ \textbf{Status}} &
    \textbf{Family Size}   &
    \textbf{Income}  &
    \makecell{\textbf{Languages} \\ \textbf{Spoken}}\\ 
    \hline
    \hline
    V3N1P5   & F      & 19   & Latin &Immigrants & No degree & Ed services &Married & 5+ & $<$\$10k & 1 \\\hline
    V3N1P5   & F      & 65+  & Chinese & Immigrants & No degree &Food services &Widowed & 3 & \$10k to \$19k & 2\\\hline
    V3N1P5   & M      & 51   & Korean & Immigrants & College &Waste management &Separated & 1 & $<$\$10k & 2   \\\hline
    V3N1P5   & M      & 60   & Korean & Immigrants & Master &Finance & Married & 2 & $<$\$10k & 2        \\\hline
    \hline
    \end{tabular}
    \end{adjustbox}
    \caption{An excerpt of simulated data for one postal region}\label{tab2}
    
    \vspace{2\baselineskip}
    \centering
        \begin{adjustbox}{max width=\textheight}
        \begin{tabular}{c|c|c|c|c|c|c|c|c|c|c|c|c|c|c}        
        \hline
        \makecell{\textbf{PostalCode}}& 
        \makecell{\textbf{Gender}}& 
        \makecell{\textbf{PersonAge}} & 
        \makecell{\textbf{old car VOI}} & 
        \makecell{\textbf{Dealer where } \\ \textbf{old car was} \\ \textbf{purchased}}& 
        \makecell{\textbf{date first} \\ \textbf{email sent}}& 
        \makecell{\textbf{date last} \\ \textbf{email sent}}& 
        \makecell{\textbf{unsubed while} \\ \textbf{in LYOL} \\ \textbf{ flag Y/N}}& 
        \makecell{\textbf{Finished Full}\\\textbf{Cadence}}& 
        \makecell{\textbf{Lease }\\ \textbf{terminated} \\\textbf{while in LYOL}}& 
        \makecell{\textbf{Sold while}\\\textbf{ in LYOL}}& 
        \makecell{\textbf{Lease renewed}\\\textbf{flag Y/N}} & 
        \makecell{\textbf{new Purchase}\\ \textbf{Date}}& 
        \makecell{\textbf{new car}\\ \textbf{signature}}& 
        \makecell{\textbf{dealer where} \\ \textbf{new car was}\textbf{purchased}}\\
        \hline \hline
        V3N1P5           & M      & 53        & XXX SUV 3   & Brian Jessel Dealership            & 4/9/2018              & 10/18/2018           & N                              & Y                     &                                &                    &                        &                   &                   &                                    \\
        H7T1T4           & M      & 68        & XXX SUV 4   & XXX Dealership in Laval            & 2/20/2018             & 6/20/2018            & N                              &                       & Y                              &                    & Y                      & 7/10/2018         & Lease             & XXX Laval                          \\

        L9X0S4           & M      & 45        & XXX Sedan 3 & Georgian XXX                       & 5/6/2019              & 7/8/2019             & N                              &                       &                                &                    &                        &                   &                   &                                    \\
        H9B1A5           & F      & 45        & XXX Sedan 3 & XXX Canbec                         & 1/17/2019             & 7/24/2019            & N                              & Y                     &                                &                    &                        &                   &                   &                                    \\
        L4S1W5           & M      & 52        & XXX SUV 3   & XXX Autohaus                       & 3/29/2018             & 10/11/2018           & N                              & Y                     &                                &                    & Y                      & 10/31/2018        & XXX SUV 3         & Lease                              \\
        H7P0B9           & F      & 43        & XXX Sedan 3 & XXX Laval                          & 4/1/2019              & 6/25/2019            & N                              &                       &                                &                    &                        &                   &                   &                                   \\
    \hline \hline
    \end{tabular}

    \end{adjustbox}
    \caption{An excerpt of auto leasing data}\label{tab3}
\end{sidewaystable}

\section{Results and Discussion}
In this section, we demonstrate the quality of SynC using two real world examples. In both cases, our coarse data comes from the 2016 Canadian National Census, which is collected by Statistics Canada and compiled once every five years. The census is aggregated at the postal code level and made available to the general public. There are 793,815 residential postal codes in Canada (in the format L\#L\#L\#, where L is a letter and \# is a digit), with an average of 47 residents per postal code. The dataset contains more than 4,000 variables ranging from demographics, spending habits, financial assets, and social values. Table \ref{tab1} illustrates a subset of this dataset with 3 postal codes and 4 variables.

Demographics variables from the Census is chosen as the core features because this subset is by far the most comprehensive portion of all parts. All Canadian residents are required to complete this section, while other sections are optional and often complemented by other sources such as private market research data (usually much smaller in sample size and coverage). 

In the following two case studies, we validate:
\begin{enumerate}
    \item the improvements on model accuracy when SynC is used as a feature engineering tool when training data is limited,
    \item the quality of SynC by comparing a sample of the generated individual data against real residents in these locations, and therefore leveraging it as an alternative to traditional market research.
\end{enumerate}

\subsection{SynC as a Feature Engineering Tool}
To assess the performance of SynC as a feature engineering tool, we collaborate with a global automotive company (hereafter referred to as the "Client") that specializes in producing high-end cars to build a predictive model to better assist their sales team in identifying which of their current customers who have a leased vehicle are interested in buying the car in the next 6 months. This type of analysis is extremely important in marketing, as contacting customers can be both expensive (e.g. hiring sales agents to make calls) and dangerous (e.g. potentially irritating and leading to unsubscriptions of emails or services). Therefore building accurate propensity models can be extremely valuable.

Our experiment contains three steps. 1) We work with the Client to identify internal sales data and relevant customer profiles such as residential postal code, age, gender, 2) we apply \textit{probabilistic matching} to join sampled data together with the Client's internal data to produce an augmented version of the training data, and 3) we run five machine learning models on both the original and the augmented data, and evaluate the effectiveness of SynC.

\subsubsection{Data Description}
There are 7,834 customers who currently lease a specific model of car from the Client in Canada. Our Client is interested in predicting who are more likely to make a purchase in the next 6 months. In Table \ref{tab3}, we attach an excerpt of the Client's sale records. For security reasons, names and emails are removed. 

To augment this dataset, the Client selects 30 variables from the Census, which included information they do not know but could be useful features. The variables includes, demographics (personal and family), financial situations and how they commute to work. We include an excerpt of the sampled individual data obtained by apply SynC to the Census in Table \ref{tab3}; both datasets are available upon request.

\subsubsection{Probabilistic Matching}
One challenge of using SynC as a feature engineering tool is the fact that synthetic population is anonymous. In most applications, enterprise level data sources are individually identifiable through an identifier which may be unique to the company (e.g. customer ID for multiple products/divisions at one company) or multiple companies (e.g. cookie IDs which can be shared between multiple apps/websites). This makes merging different data sources very easy as the identifier can be used as the primary key. By construct, however, synthetic population individuals are model generated from anonymous data, and therefore cannot be joined by traditional means. Here we present an alternative method called \textit{Probabilistic Matching}.

Because SynC produces anonymous information (i.e. data that cannot be attributed to a specific person, such as name, email address or ID), merging of variables can only be done via variables that are indicative but not deterministic. Age, gender, ethnicity, and profession are good examples of such variables. In table 3 we show the first few customers supplied by our industry parnter. We also provide the list of synthetically generated persons for postal code V3N1P5 in Table \ref{tab2}, and use this as an example to demonstrate how probabilistic matching can be done on the first customer.

Client data show that a 53 year old male, who lives in the area of V3N1P5 made a lease purchase. In this case, we have three indicative measurement for this customer - this buyer is \textit{53 years old}, \textit{male} and lives in \textit{V3N1P5} (an area in Burnaby, British Columbia, Canada). In our synthetic data, the closet match would be the tenth person, as two of the three indicators (postal code and genders) match precisely, and the last indicators matches closely (age of 54 vs. 53, which is a difference of 1 year). We can conclude that this customer, who leased an SUV of model type 3, is likely to be ethnically Chinese, an immigrant with a bachelor's degree, an income of between \$90k to \$99k and speaks 2 languages.

Notably, the third person is also a very close match. Both postal code and gender matches perfectly, and age matches closely as well (51 vs.\  53, which is a difference of 2 years rather than 1 year). It would also make sense to set a threshold (such as within 5 years difference in age), and aggregate all matched records via averaging (for numerical variables) or voting/take proportions (for categorical variables).

\begin{table}[!t]
    \resizebox{\columnwidth}{!}{
    \begin{tabular}{c|c|c|c|c|c}
    \hline
     & \textbf{LR} & \textbf{DT} & \textbf{RF} & \textbf{SVM} & \textbf{NN} \\ 
     \hline \hline
    Original Data & 61.5\% & 63.9\% & 70.4\% & 69.3\% & 68.8\% \\ 
    \hline
    Augmented Data & 66.2\% & 71.1\% & 73.0\% & 80.6\% & 73.9\% \\
    \hline 
    \end{tabular}}
    \caption{Comparisons of accuracy measures of 5 different classifiers trained on original data and synthetic population augmented data}
    \label{performance_classifier}
\end{table}

\subsubsection{Method Evaluation}
We train 5 different classifiers on both the partner's data, as well as the augmented dataset to predict whether a customer buys the leased vehicle. We train Logistic Regression (LR), Decision Tree (DT), Random Forest with 500 tress (RF), SVM with Radial Basis Kernel and 2-Layer Neural Network (NN), and their performances can be found in Table \ref{performance_classifier}. 

In all five cases, augmented data produces a higher classification accuracy than the raw data from our industry parnter. Accuracy increases range from slightly over 2.5\% (RF) to as much as 11\% (SVM), with an average increase of 6.2\%. This increase is both technically significant, as well as practically meaningful, as the Client would easily apply this model to their business and achieve grow their sales. 

This case study has shown that Synthetic Population is an effective way to engineer additional features to situations where the original training data is limited. As explained in early sections, SynC takes coarse datasets and generate estimates of individuals that are likely to reside within the given postal code area. Although SynC does not produce real people, the generated "synthetic" residents both closely resembles the behaviours of true population and is also consistent with the available sources. We demonstrate that it is a viable data augmentation technique.

\subsection{SynC as an Alternative to Market Research}
The validity is hard to assess due to lack of true individual data to benchmark. After all, if individual data was attainable then SynC would not be needed in the first place. Therefore, We propose an alternative evaluation metric which requires 1) a set of surveyed response representing the population and 2) simulated populations from the postal region of the surveyed individuals. 
For a surveyed dataset containing $D$ features from $T$ individuals across $M$ postal codes (each containing $n_m$ surveyed individuals), $M$ simulated populations each with $T_m$ (the population of $m^{th}$ postal region) individuals are generated using the proposed framework. 

The evaluation function, $\mathcal{L}$ is defined as the following:
\small
\begin{equation}
    \mathcal{L} = \frac{1}{M}\sum_{m=1}^{M}\frac{1}{n_m}\sum_{k=1}^{n_m}\frac{1}{D}\sum_{d=1}^{D} \mathbbm{I}\{x_{m, k}^d = y_{m, k}^d\}
\end{equation}
\normalsize
and 
\small
\begin{equation}
    \{y_{m,k}\}^{n_m}_{k=1} = \argmax_{\{y_{m,k}\}_1^{n_m}} \sum_{k=1}^{n_m}\sum_{d=1}^{D} L(x_{m,k}^d, y_{m,k}^d)
\end{equation}
\normalsize
where
\small
\begin{equation}
L(x, y) =\left\{
    \begin{array}{@{}ll@{}}
    \mathbbm{I}\{x_{m,k}^d = y_{m,k}^d\}, & \text{if $x_{m,k}^d$ is categorical} \\
    \frac{|x_{m,k}^d - y_{m,k}^d|}{x_{m,k}^d}, & \text{if $x_{m,k}^d$ is continuous}
    \end{array}\right.
\end{equation}
\normalsize
Intuitively, $\mathcal{L}$ measures the average similarity across all surveyed postal codes between the person from true population and the set of simulated individuals in each region that resembles them the most. 

We collect responses from 30 individuals in Toronto, Canada. For simplicity, the surveyed questions only contain five core features (Age, Sex, Ethnicity, Education, Income), seven additional personal features (Immigration Status, Marital Status, Family Size, Profession etc.) and four spending behaviour feature (Favourite Store for Cloth/Food/Grocery/Furniture). The five demographics features are selected as core variables and the others are grouped into two batches, personal and spending behaviour. Using the SynC framework along with a 2-layered neural network in \textbf{Phase III}, a sample is produced and an excerpt is shown in Table \ref{tab2}.

Table \ref{tab3} presents three comparison pairs between a few surveyed individuals and their simulated counterparts (pairs are highlighted by color). With this sample, $\mathcal{L} =  82\%$. On a closer look it is easy to see that the generated individuals are more realistic than what the score can reflect. The metric function uses an indicator function for each feature to evaluate the accuracy, which does not capture the potential distance among categories (e.g. $Income_{50k-59k}$ should be ``closer" to $Income_{60k-69k}$ compare to $Income_{150k+}$). 

\begin{sidewaystable}
\vspace*{10pt}
\begin{adjustbox}{max width=\textheight}
\begin{tabular}{c|c|c|c|c|c|c|c|c|c|c}
\hline
\textbf{Postal} &  \textbf{Sex} & \textbf{Age} & \textbf{Ethnicity} &\makecell{\textbf{Immigration} \\ \textbf{Status}}  &\textbf{Education} &\textbf{Profession} &\makecell{\textbf{Marital} \\ \textbf{Status}} &\textbf{Family Size}   &\textbf{Income}  &\makecell{\textbf{Favorite} \\ \textbf{Store}}\\ 
\hline
\hline
M2M4L9   & M      & 60   & Korean & Immigrants & Master &Finance & Married & 2 & $<$\$10k & H\&M        \\\hline
M2M4L9   & F      & 58   & Other & Non-immigrants & College &Health care & Married        & 3 & \$50k to \$59k & Hot Renfrew \\\hline
M2M4L9   & M      & 26   & Chinese & Immigrants &HighSchool&Retail & Never married  & 3  & $<$\$10k & Old Navy    \\\hline
M2M4L9   & M      & $<$14 &Mid Eastern & Immigrants & N/A & N/A & N/A & N/A & N/A & N/A \\\hline
M2M4L9   & F      & 21   & Chinese & Immigrants & Bachelor & Retail & Never married  & 3 & <\$10k & Gap \\\hline
M2M4L9   & M      & 51   & Chinese & Immigrants & No degree & Other services & Never married  & 2 & \$10k to \$19k & Gap\\ \hline
\end{tabular}
\end{adjustbox}
\caption{An expert of simulated data for one postal region}\label{tab3}
\vspace*{20pt}

\vspace*{10pt}
\begin{adjustbox}{max width=\textheight}
\begin{tabular}{c|c|c|c|c|c|c|c|c|c|c|c}
\hline
\textbf{Type} &\textbf{Postal} &  \textbf{Age} & \textbf{Sex} & \textbf{Education} & \textbf{Ethnicity} &\textbf{Family Size}  &\makecell{\textbf{Immigration} \\ \textbf{Status}} &\makecell{\textbf{Marital} \\ \textbf{Status}} &\textbf{Income} &\textbf{Profession} &\makecell{\textbf{Favorite} \\ \textbf{Store}} \\ 
\hline
\hline
\rowcolor{beaublue}
Simulated	&M2M4L9	&47 &F &Bachelor &Chinese &2 &Immigrants &Married &\$40k to \$49k	 &Scientific/Technical &Gap\\ \hline
\rowcolor{beaublue}
Surveyed	&M2M4L9 &49 &F &Bachelor &Chinese &3 &Immigrants &Married &\$5k to \$59k	&Scientific/Technical &H\&M\\ 
\hline
\hline
\rowcolor{beauyellow}
Simulated	&M2M4L9	&46 &M &Bachelor &Chinese &2 &Immigrants &Married &$<$10k	 &Scientific/Technical &Gap\\ \hline
\rowcolor{beauyellow}
Surveyed	&M2M4L9 &51 &M &Bachelor &Chinese &3 &Immigrants &Married &\$5k to \$59k	&Scientific/Technical &Gap\\ 
\hline
\hline
\rowcolor{beaublue}
Simulated	&M4Y1G3 &22 &M &Bachelor &Filipino &3 &Immigrants &Married &\$70k to \$79k &Scientific/Technical &Banana Rep\\ \hline
\rowcolor{beaublue}
Surveyed	&M4Y1G3 &23 &M &Bachelor &Chinese &2 &Immigrants &Never married &\$5k to \$59k	&Finance &Banana Rep\\ 
\hline
\end{tabular}
\end{adjustbox}
\caption{An comparison between simulated samples and surveyed samples}\label{tab4}
\end{sidewaystable}
\section{Conclusion and Future Directions}
In this work, we propose a novel framework, SynC, for generating individual level data from aggregated data sources, using state-of-the-art machine learning and statistical methods. To show the proposed framework's effectiveness and boost reproducibility, we provide the code and data on the Canada National Census example described in \textit{Proposed Method} Section. We also present a real-world business use case to demonstrate its data augmentation capabilities. 

As a first attempt to formalize the problem, we see three areas where future works can improve upon. First of all, our method relies on Gaussian copulae and this can be further extended by leveraging other families of copulae to better model the underlying dependency structures. Secondly, we use beta and log-normal distributions to approximate marginal distributions for categorical and continuous variables, respectively, and other families of distributions could be considered (e.g., the $\kappa$-generalized model \shortcite{clementi2016kappa} can be used for money related distributions). Lastly, including more geographical features can potentially increase the accuracy of predicted individual level traits.

%
%
\bibliographystyle{aaai}
\bibliography{mybibliography}

\end{document}